\documentclass{IOS-Book-Article}
\usepackage{times}
\normalfont
\usepackage{url}
\usepackage[T1]{fontenc}
\usepackage{amsmath}
\usepackage{amsthm}

\usepackage{multirow}
\usepackage{epsfig}

\begin{document}

\begin{frontmatter}                           

\title{Towards quantitative measures in applied ontology}
\runningtitle{Quantitative measures}

\author[A]{Robert Hoehndorf\thanks{Corresponding Author: Robert
    Hoehndorf, rh497@cam.ac.uk.}}
\author[B]{Michel Dumontier}
\author[A]{Georgios V. Gkoutos}

\runningauthor{Hoehndorf, Dumontier \& Gkoutos}
\address[A]{Department of Genetics, University of Cambridge}
\address[B]{Department of Biology, Institute of Biochemistry and School
    of Computer Science, Carleton University}

\begin{abstract}
  Applied ontology is a relatively new field which aims to apply
  theories and methods from diverse disciplines such as philosophy,
  cognitive science, linguistics and formal logics to perform or
  improve domain-specific tasks. To support the development of
  effective research methodologies for applied ontology, we critically
  discuss the question how its research results should be
  evaluated. We propose that results in applied ontology must be
  evaluated within their domain of application, based on some
  ontology-based task within the domain, and discuss quantitative
  measures which would facilitate the objective evaluation and
  comparison of research results in applied ontology.
\end{abstract}

\end{frontmatter}

\thispagestyle{empty}
\pagestyle{empty}

\section{Introduction}
Applied ontology is an emerging discipline that applies theories and
methods from diverse disciplines such as philosophy, cognitive
science, linguistics and formal logics to perform or improve
domain-specific tasks. Scientific disciplines require a research
methodology which yields reproducible and comparable results that can
be evaluated independently. Methodological progress in applied
ontology will be recognized when different methods generate results
that can be objectively compared, such that it can be evaluated as to
whether the methods yield better results. To illustrate, {\em text
  mining}, which was firmly established as a scientific discipline in
the early 1990s, leveraged knowledge from computational linguistics,
cognitive science, philosophy, graph theory, machine learning, and
other areas of computer science to create new, large scale methods to
uncover information from natural language documents. Although text
mining could, in principle, be evaluated from the perspective of
computational linguistics (i.e., how well a particular linguistic
theory was implemented and applied), it is most commonly evaluated
from the perspective of scientific contribution (i.e., how well some
scientific question was addressed by the text mining method) through
quantitative measures that include precision and recall based on
comparison to a gold standard, the F-measure (i.e., the harmonic mean
between precision and recall), the area-under-curve (AUC) in an
analysis of the receiver operating characteristic (ROC) curve
\cite{Fawcett2006}, or the use of kappa-statistics to determine
agreement between manual evaluation by domain experts. Establishment
of a common set of (quantitative) measures has made it possible to
compare different methods and approaches in text mining with respect
to their contribution to particular tasks and has the field as a whole
allowed to measure its progress.

Research in applied ontology currently lacks established quantitative
metrics for evaluating its results. More importantly, applied ontology
lacks agreement about the perspective from which its results should be
evaluated.  Evaluation of applied ontology research is more often than
not based on criteria stemmed from philosophy, knowledge
representation, formal logics or ``common sense'', while an evaluation
based on the domain of application is rarely performed
\cite{Obrst2007}.  In the absence of commonly agreed criteria for
evaluating research results, the evaluation and comparison of research
in applied ontology is prone to subjectivity, lack of transparency,
opinion, preference and dogma. Furthermore, the lack of established
evaluation criteria for applied ontology research hinders the
development of an effective research methodology for the field of
applied ontology: before a research methodology in any scientific
field can be established, it is first necessary to determine what
constitutes a research result, what constitutes a {\em novel} research
result (i.e., what does it mean that two research results are
different), and what constitutes a better result than another (i.e.,
how can two competing results be compared and evaluated). Only after
these questions are answered will it be possible to design a research
methodology in a scientific field than enables the field as a whole to
make progress with respect to the evaluation criteria that the
discipline has established.

Here, we being to explore ways for evaluating research in applied
ontology.  Our arguments and examples will primarily focus on
ontologies that are used in science, in particular for biomedicine,
but we believe that many of our arguments will hold for research in
other areas of applied ontology as well. From a certain perspective,
this work is a continuation and extension of the work of Barry Smith
\cite{Smith2008}, who made one of the first moves towards questioning
the status of {\em applied ontology} as a scientific discipline, and
who already stated that ``[c]entral to ontology (science) is the
requirement that ontologies [...] should be tested empirically''.

Our central claim will be that research in applied ontology must be
evaluated within the domain to which it is being applied. More
precisely, we claim that research results in applied ontology need to
be evaluated with respect to a {\em specific task} that is supposed to
be achieved, and that any contribution in applied ontology should be
motivated by a task and evaluated based on it.  In many cases, the
results of such an evaluation can be {\em quantified}, leading to
objective, empirical criteria for the success of applying ontological
methods within a domain.  We do not intend to propose a comprehensive
list of such evaluation criteria, but rather aim to make suggestions
for the research community in the hope to spawn a discussion that
results in community agreement and a first step towards the
development of a commonly agreed research methodology for our field.

\section{Biomedical ontologies}
At the end of the 1990s and early 2000s, genetics made a leap forward
with the availability of the first genome sequences for several
species. The availability of genome sequences for multiple species
enabled comparative genomic analysis, and it was recognized that a
large part of the genetic material in different species was conserved
and that many of the genes in different organisms have similar
functions. The Gene Ontology (GO) \cite{Ashburner2000short} was
designed as a controlled vocabulary to provide stable names, textual
definitions and identifiers to unify descriptions of functions,
processes and cellular components across databases in biology.  Today,
with the rise of high-throughput sequencing technology, genome
sequences for thousands of species are becoming available, and large
international research projects, such as the 5,000 genomes project
(which aims to sequence the genomes of 5,000 insects and other
arthropods) or the Genomes 10k project (which aims to sequence the
genomes of 10,000 vertebrate species), will collect even more data in
the near future.  High-throughput technologies are not limited to
genome sequencing, but entered other areas in biology as well, from
high-throughput phenotyping (to determine the observable
characteristics of organisms, often resulting from targeted mutations)
over microarray experiments (to determine gene expression) to
high-throughput screening (in drug discovery). The amount of data
produced in biology today makes the design of strategies for
integration of data across databases, methods for retrieving the data
and developing query languages and interfaces a central and important
part of research in biology.  The prime purpose of ontologies such as
the GO is to address these challenges that arose in biology and
bio-medicine within the last few years.

To facilitate the integration of databases, retrieval of data and the
provision of query languages, ontologies provided not only terms and
textual definitions, but also a basic structure. Initially, this
structure was not expressed in a formal language (i.e., a language
with an explicit semantics). Instead, ontologies were seen as graph
structures in which nodes represent terms and edges relations (such as
{\bf is-a} or {\bf part-of}) between them. Reasoning over these graphs
was stated as operations on the graph, in particular the composition
of edges and the transitive closure \cite{Ashburner2000short}. It was not
until much later that formal languages were used to represent
biomedical ontologies and recast the graph operation in terms of
deductive inference over formal theories \cite{Horrocks2007,
  Golbreich2007, Smith2005, Hoehndorf2010patterns}.

The graph structure of biomedical ontologies is not only a valuable
feature to improve retrieval and querying, but is widely used in the
form of Gene Set Enrichment Analysis (GSEA) \cite{Subramanian2005} to
analyze gene expression. GSEA utilizes the graph structure of the GO
to determine whether a defined set of genes shows statistically
significant, concordant differences between two biological states; it
utilizes the annotation of sets of genes with GO terms and the GO
graph structure and inference rules to statistically test for enriched
GO terms. A large number of tools were developed to perform GSEA, and
it has lead to discoveries of cancer mechanisms
\cite{Subramanian2005}, evolutionary differences in primates
\cite{Prufer2007}, and GSEA is now a standard tool in many biological
analyses, as evidenced by more than 3,000 citations\footnote{Based on
  Google Scholar, 12 Jan 2012.} for the original paper. Similar
enrichment analyses are now being performed using ontologies of other
domains, such as the Human Disease Ontology \cite{LePendu2011}.

The graph structure of ontologies is also widely utilized for semantic
similarity analyses \cite{Resnik1999}. Semantic similarity measures
apply a metric on an ontology in order to compare the similarity
between data that is annotated with classes in the ontology. Some
metrics are based on the distance between two nodes in the ontologies'
graph structure, while others compare sets of classes that are closed
with respect to relations in the ontology. In some cases, the metrics
include further information, such as the information content that a
class in an ontology has within a given domain. Importantly, however,
semantic similarity measures rely on the number and the kind of
distinctions that the ontology developers have made explicit, and
utilize the {\em explicit} semantics in an ontology only indirectly.

Another application of ontologies is in text mining and literature
search and retrieval. The availability of a common terminology
throughout biology enables the task of named entity recognition, i.e.,
the identification of standardized terms in natural language
texts. When terms from ontologies can reliable be detected in natural
language texts, ontologies can be used for retrieving text documents
from literature archives such as PubMed \cite{Doms2005}. This task is
made easier when terms in ontologies are widely used, and several
biomedical ontologies have been evaluated based on how well their
terms can be recognized in scientific literature \cite{Yao2011}.

Neither of the applications of biomedical ontologies we discussed so
far actually relies on formalized semantics, axioms, the use of
knowledge representation languages and automated reasoning, or
philosophical foundations. Nevertheless, the past years have seen a
rapid increase in applying formal methods to biomedical ontologies. In
particular, the Web Ontology Language (OWL) \cite{Grau2008} is now
widely used to represent biomedical ontologies \cite{Horrocks2007}. In
some cases, more expressive languages such as first- and monadic
second-order logic is used to specify ontologies, in particular for
biological sequences \cite{Hoehndorf2009sequences} and molecular
structures and graphs \cite{Hastings2011}.  The stated aims of using
the axiomatic method \cite{Hilbert1918} and knowledge representation
languages for biomedical ontologies are manifold, and include, among
others, the search for philosophical rigour and a foundation in
``good'' philosophy \cite{Hastings2010}, providing ``unambiguous''
documentation of the meaning of terms in a vocabulary \cite{rnao},
verifying the consistency of a (conceptual) data model
\cite{biopaxshort, Mungall2010go}, verifying the consistency of data
with respect to a data model \cite{biopaxshort, Hoehndorf2011models},
enabling complex retrieval and querying through automated reasoning
\cite{Ruttenberg2007short}, integrating multiple ontologies
\cite{Hoehndorf2011incon, Mungall2011}, and decreasing the cost of
developing and maintaining an ontology \cite{goble,
  Bada2004}. Furthermore, the application of formal methods in
biomedical ontologies has the potential to reveal mistakes in the
design of ontologies and thereby improve their utility for scientific
analyses \cite{Smith2005, Smith2003}. Several projects have started to
axiomatize biomedical ontologies \cite{Mungall2010go, Mungall2011,
  Eilbeck2010, wpo}, and these projects have led to changes in the
ontologies and the detection and removal of contradictory statements
\cite{Hoehndorf2011incon, Mungall2010go}.  Other researchers have
suggested changes to improve ontologies' structures and axioms based
on applying formal, ontological and philosophical methods
\cite{Smith2005, Smith2003, Schulz2009, Hoehndorf2010}, or they
provide ontological interpretations of domain-specific knowledge by
applying some formal ontological theory to some phenomena in a domain
\cite{Rohl2011, Schulz2009, el1, Schulz2008}.

Despite the large number of research projects that apply formal
ontological theories to (scientific) domains, no common evaluation
criteria are being applied in these studies. Examples of criteria of
evaluation include formal consistency \cite{Kutz2011}, identification
of unsatisfiable classes \cite{Mungall2010go, Hoehndorf2011incon},
conformance to a ``good'' philosophy (i.e., some particular
philosophical view) \cite{Smith2005wuesteria, Smith2010realism,
  Hastings2010}, user acceptance \cite{Boeker2011}, conformance to
naming conventions \cite{Schober2009} or the recall of ontology class
labels in scientific literature \cite{Yao2011}. Only few of these
criteria actually evaluate the {\em application} of ontologies to some
task, while the majority of these criteria evaluate the research
results based on philosophical, formal and technical criteria that lie
within the domain of ontology or its underlying technologies
themselves.

\section{Towards quantitative evaluation criteria for research results
  in applied ontology}
The selection and application of evaluation criteria may provide us
with the means to distinguish research in {\em applied} ontology from
research in non-applied ontology. In {\em applied} ontology,
ontologies are being used for some task within a domain, and that task
lies usually outside of the domain of ontology itself\footnote{A
  notable exception to this is when we apply ontological methods to
  the domain of ontology itself, and classify different kinds of
  ontology, analyze the types of relations between classes, relations,
  instances and individuals, etc. Such an ontology could, for example,
  be used to provide the conceptual foundation of an ontology editor,
  to enable interoperability between different ontology learning
  algorithms, in portals providing access to different ontologies, or
  in an ontology evaluation framework.}. Consequently, quality
criteria for research in applied ontology will be derived from the
task to which the research is being applied, and not from the domain
of ontology itself. On the other hand, the search for philosophical
foundation and rigour, including the demonstration {\em that} a
particular philosophical theory is capable of expressing distinctions
that are being made within a domain, are examples of research goals of
non-applied ontology, not of applied ontology, because the {\em aim}
of the research and its evaluation will generally lie within the realm
of ontology or philosophy, not within the domain of application.
Applying a particular philosophical theory can, in many cases, improve
the utility of an ontology within a domain. Nevertheless, the fact
{\em that} a philosophical theory can be applied within a domain alone
does not, in our opinion, constitute a result in applied ontology; on
the other hand, that the application of a particular philosophical
theory or perspective {\em improves} the utility of an ontology for
some task in a domain would constitute a result in applied ontology.

We can also observe {\em who} or {\em what} directly benefits from a
particular aim of research in ontology: either the users and uses of
an ontology, ontology-based applications, and specific tasks to which
ontologies are being applied, or the developers and maintainers of an
ontology. Developers and maintainers of ontologies will benefit
directly from decreased maintenance work, ease of construction and the
availability of technical documentation, while users and applications
of an ontology will only benefit indirectly from such research goals
(and these benefits would normally have to be demonstrated).  Users
and applications of ontologies benefit from the community agreement
which ontologies can bring about and their resulting potential for
ontology-based data annotation and integration, retrieval and
querying, novel scientific analyses, and in some cases consistency
verification of data. In particular, users and uses of ontologies will
benefit from something that ontologies can {\em do}, and research in
{\em applied} ontology -- ontology research to serve some domain's use
case -- will have to be measured on how well they perform their task.

One of the most widely cited applications of ontologies in science is
their potential to facilitate community agreement of the meaning of
terms in a domain. These terms are frequently used as metadata in
scientific databases and publications. Consequently, applying
ontologies to standardize the vocabulary used as meta-data can enable
the integration and interoperability of databases and research
results. Yet, how could such a research result -- an ontology that is
intended to effectively standardize the meaning of terms in a
vocabulary in order to support interoperability and integration -- be
evaluated? Since the prime aim of such a research result is to achieve
community agreement, an obvious evaluation criterion would be to
conduct a user-study that evaluates whether different users can
consistently apply terms within a standardized task such as the
annotation of a data set with classes from an ontology. For this task,
Kappa statistics can be applied and a $\kappa$ value can be reported
that measures the degree to which annotator can consistently apply an
ontology within the task \cite{Cohen1960}. Alternatively, an
integrated scientific analysis of the data in multiple databases
between which interoperability is intended to achieve can be performed
and evaluated on a scientific use case. For example, the development
of formal definitions for phenotype ontologies \cite{Mungall2010}
could be quantitatively evaluated by using these definitions to
integrate multiple model organism databases and analyze the integrated
knowledge with regard to its potential for revealing novel candidate
genes for diseases \cite{Hoehndorf2011phenome}.

The support of queries and the accurate retrieval of data is another
task that ontologies or the axioms in ontologies are developed for.
Information retrieval is a discipline in computer science for which
rigorous quantitative evaluation criteria are available
\cite{VanRijsbergen1979}, often based on the comparison to a gold
standard or a set of positive and negative examples based on which
statistical measures can be applied. Quantitative measures include the
F-measure (the harmonic mean between precision and recall) or the
area-under-curve in an analysis of the receiver operating
characteristic (ROC) curve \cite{Fawcett2006}. If an ontology, or
axioms in an ontology, are intended for retrieval, measures of this
kind can be applied to demonstrate the success.

In many cases, axioms in ontologies are added in order to enable novel
queries that make distinctions which could not be made before. For
example, adding axioms that assert a partonomy to a purely taxonomic
representation of anatomical structures {\em enables} new kind of
queries based on the use of parthood relations. Such a result -- the
addition of new axioms in order to enable novel types of queries and
retrieval operations -- can be evaluated using the same quantitative
measures as ontology-based retrieval.  All of these descriptions
assume that there is already some data which is being retrieved using
queries over the ontology. In the absence of such data, e.g., when a
new ontology is proposed within a domain with the intent to use this
ontology to annotate data in the future, data could still be simulated
and used in the evaluation.

Further applications of formalized ontologies include the verification
of data with respect to certain constraints that are expressed within
the ontology. For example, in the domain of biological
pathways\footnote{A biological pathway is a series of interactions
  that lead to a particular outcome, such as a chemical product or the
  realization of a particular function.}, the BioPax ontology
\cite{biopaxshort} has been proposed, and one of its aims is to verify
pathway data with respect to the model that the BioPax ontology
provides. Similarly, a recent study used formal ontological analysis
and automated reasoning to investigate the consistency of data stored
in the BioModels database (a database of computational models in
systems biology), and identified a large number of incorrectly
characterized database entries \cite{Hoehndorf2011models}.

Applications of ontology research in scientific analyses and in the
process of making novel scientific discoveries are maybe the best
evaluated contributions in applied ontology, since the contributions
that ontology research can make in these areas is commonly subject to
the same evaluation criteria as other contributions in the scientific
domain of application. For example, the GSEA method was evaluated both
using statistical measures and experimentally verified data that has
been extensively studied \cite{Subramanian2005}, and the use of
semantic similarity measures to identify interacting proteins based on
similar Gene Ontology annotations is rigorously evaluated and compared
using ROC analysis and correlation coefficient analysis
\cite{xu2008}. In each case, the scientific domain to which
ontology-based methods are being applied has established (and often
demands) quantitative evaluation criteria that can guarantee -- at
least to some degree -- the objective and empirical evaluation and
comparison of research results.

There are several other tasks that may fall in the domain of applied
ontology research. For example, formal ontological analysis can be
applied to specify a (conceptual) model, verify its consistency and
identify modelling choices that potentially lead to faulty results; or
formal ontology can be applied to formally and ``unambiguously''
specify the meaning of terms in a vocabulary (e.g., to enable
communication between autonomous intelligent agents). Some of these
tasks can also be evaluated quantitatively: while consistency of a
conceptual model is a binary quality that relies on a consistency
proof, incorrect consequences can be estimated using predefined tests
that aim to make inferences of a certain kind. The ``unambiguous''
formal specification of the meaning of a term using an ontology would
require a meta-theoretical analysis and a completeness proof for the
ontology.

\begin{table*}
\centering
\begin{tabular}{|p{4cm}|p{4cm}|p{3cm}|}
  \hline
  Application & Potential evaluation method & Quantifiable result\\
  \hline&&\\
  Establish community agreement of the meaning of terms within a
  domain; facilitate data annotation; support integration and
  interoperability & User study; integrated analysis; completeness
  proof & Inter-annotator agreement, Cohen's Kappa \\
  \hline
  Retrieval & Comparison to gold standard, ROC analysis, unit tests &
  area-under-curve, F-measure, precision, recall\\
  \hline
  Scientific analysis (GSEA, semantic similarity) & Comparison to gold
  standard, statistical analysis & p-value, area-under-curve,
  F-measure, precision, recall\\
  \hline
  Consistency of data & automated reasoning, performance evaluation &
  computational complexity\\
  \hline
  Determine the consistency of a (conceptual) model & automated
  reasoning, consistency proof & consistent or not (binary)\\
  \hline
  Test the accuracy of a (conceptual) model & automated reasoning, unit tests
  (for inferences), unit tests (for application) & number of
  unsatisfiable classes, number of tests passed/failed\\
  \hline
\end{tabular}
\caption{\label{tbl:ontologies}Opportunities for the quantitative
  evaluation of research results in applied ontology.}
\end{table*}


To summarize, depending on the task that is being performed using some
ontology research result, we will be able to derive different quality
criteria, some of which are illustrated in Table
\ref{tbl:ontologies}. However, the heterogeneity of ontology-based
applications and ontology-driven approaches prevents the application
of a single quality and evaluation criteria. Instead, we have to
evaluate research results in applied ontology in conjunction with a
particular task to which this result is being applied. For example,
instead of evaluating the quality of an ontology $O$ that represents
biological pathways, we evaluate $O$ with respect to different tasks
that it is intended to perform. For example, $O$ may be used to
achieve community agreement about the terms used to annotate pathway
databases (task $t_1$), and we can evaluate $O$ with respect to this
task. On the other hand, $O$ may also be used to verify the
consistency of biological pathway data (task $t_2$), and we may
evaluate $O$ with respect to $t_2$. It may then turn out that $O$
achieves one task very well while its performance in a second task is
poor.

Finally, robustness of research results in applied ontology can be
evaluated based on how well a research result in applied ontology
performs in multiple tasks, or how well it can be adapted to other
tasks, including tasks that are performed in other domains. Robustness
can be evaluated based on how much the quantitative evaluation changes
under changing application conditions.  For example, if the research
result is an ontology that is being developed for the semantic
annotation of a particular database and has been demonstrated (e.g.,
based on a user-study and the report of the inter-annotator agreement)
to perform well in this task, changing the database and performing a
similar study and quantitative evaluation allows to evaluate
robustness: does the quantitative evaluation result change
significantly, or does it remain the same? If the quantitative
evaluation results do not change significantly under changing
conditions of application (or they improve), evidence for a {\em
  robust} research result have been found. Notably, it is the
application of quantitative evaluation criteria that enables the
direct comparability of the suitability of a research result in
different tasks, and therefore enables an objective demonstration of
robustness.

\section{Proposed evaluation and quality criteria}
Several evaluation methods for research in applied ontology have been
proposed, and multiple studies have attempted to evaluate the quality
of ontologies in several domains. To the best of our knowledge, no
study has yet emphasized the need for objective, quantitative
evaluation criteria for applied ontology research; on the contrary,
many criteria that aim to measure the ``quality'' of ontologies are
derived from philosophical considerations or based on social
considerations. In particular, several studies emphasize the need to
treat ontologies similarly to scientific publications and propose an
evaluation strategy similar to scientific peer review.  For example,
Obrst et al.\ aim to identify ``meaningful, theoretically grounded
units of measure in [ontology]'' and perform an extensive review of
previous ontology evaluation attempts, including a brief discussion of
application-based evaluation approaches and quantifiable results
\cite{Obrst2007}. However, Obrst et al.\ dismiss application-based
evaluation strategies since they are ``expensive to carry out'', and
seem to favour evaluation by humans based on principles derived from
common sense, from formal logics or from philosophy (especially in the
form of philosophical realism). A similar route is being taken by
Smith who suggests that peer review of ontologies should become
standard practice, since ``[p]eer review provides an impetus to the
improvement of scientific knowledge over time'' \cite{Smith2008}. Such
a peer review system is intended to be adopted by the OBO Foundry
ontology community \cite{Smith2007short, Smith2008}.  The use of
expert peer review to evaluate ontologies seems like an
uncontroversial suggestion since peer review is the established method
for evaluating contributions throughout science. However, different
from applied ontology, most scientific fields have widely accepted,
and in many cases objective, quantitative, criteria on which peer
reviewers can base their judgement.  The criteria for peer review
proposed by Smith \cite{Smith2008}, Orbst et al.\ \cite{Obrst2007},
and others \cite{obmlcommandments}, are largely derived from ``common
sense'' or particular philosophical positions and have not been {\em
  demonstrated} to improve the performance of ontology-based research
in any application.  Peer review cannot be used to evaluate research
results when there is no agreement as to how a discipline is supposed
to achieve scientific progress and how these achievements can be
measured. In the absence of accepted and empirically tested criteria,
peer review will merely reflect the personal opinions of the
reviewers, and not lead to a fair evaluation of a research result's
quality or its fitness for a particular purpose.

A prime example of a conflict resulting from the lack of accepted,
empirically tested evaluation criteria is the realism debate
\cite{Merill2010, foispaper, Smith2010realism}. The realism debate is
an argument between the proponent of the ``realist methodology'', who
argue that ontologies must be evaluated with respect to some form of
philosophical realism \cite{Smith2010realism, Obrst2007,
  obmlcommandments}, and researchers in applied ontology who argue for
a research methodology in which ontological decisions are motivated
and evaluated by applications and not philosophical considerations
\cite{Merill2010, foispaper}.
The difference between a philosophical and an application-centric
perspective may be one of the reasons for misunderstandings between
the two sides in this debate: while one side attacks the other in the
realm of philosophy -- where {\em philosophical} positions are
attacked and defended, and some of the arguments have been exchanged
between philosophers for thousands of years --, the proponents of an
application-centric view would expect it to be a matter of empirical
investigation to determine which ontological design decisions address
the needs of the ontology users better than another.  In many cases,
it may turn our that two philosophical theories are {\em
  indistinguishable} for a particular scientific task (e.g., when two
theories are {\em empirically equivalent}), in which case the
particular choice of philosophical explanation will not affect the
performance of an application: when it is in principle impossible to
design an experiment that can distinguish between two alternative
theories, we would leave the realm of empirical science if we attempt
to defend or attack either theory.

There are some notable previous studies which applied quantitative
measures for formalized ontologies in biomedical applications. For
example, Boeker et al.\ \cite{Boeker2011} ``aim to analyze the
correctness of the use of logic by the OBO Foundry or close-to OBO
Foundry ontologies and related mappings'', and they identify
approximately 23\% of the axioms in the evaluated ontologies as
incorrect based on the judgement of four experts. These results are
consistent with another study by Hoehndorf et al.\
\cite{Hoehndorf2011incon} that evaluates contradictory class
definitions in OBO ontologies and identifies several thousand
unsatisfiable classes using automated reasoning. Common to these two
studies is that they evaluate ontologies based on aspects that can be
derived from their formal representations alone, assuming that
considerations such as the consistency of an ontology or the absence
of undesired inferences from an ontology will always give some
indication about an ontology's quality. However, for some tasks, not
even consistency of an ontology is required. Boeker et al.\
``hypothesize that the main and only reason why [the problematic
axioms have] little affected the usefulness of these ontologies up to
now is due to their predominant use as controlled vocabularies rather
than as computable ontologies'' \cite{Boeker2011}, already
acknowledging that their evaluation has not addressed the main task
for which the evaluated ontologies are being applied, but rather some
task (retrieval through automated reasoning) that these ontologies
could potentially also be used for. Similarly, Hoehndorf et al.\
identify several unsatisfiable classes in biomedical ontologies, but
fail to identify the problems that these may cause -- except again in
the hypothetical task of using automated reasoning to answer queries
over the ontologies. Even more problematically, the evaluated
ontologies are successfully being used for automated reasoning {\em
  although} they contain unsatisfiable classes and may lead to
undesirable inferences.  In these reasoning tasks, applications such
as database queries restrict the types of operations that are being
performed over the ontologies. From a certain perspective,
applications provide an {\em interface} to formal ontologies that may
limit the ontology to a lower expressivity than the knowledge
representation language in which the ontology is formulated. For
example, if an inference mechanism that lacks the capability to
interpret negation is used to process an ontology, retrieval
operations can be successful even in the presence of contradictory
class definitions or inconsistencies. Similarly, undesired inferences
may disappear when only certain kinds of queries can being performed.
In some cases, ``incorrect'' consequences may even be desirable: for
example, in analyses that utilize measures of semantic similarity, the
similarity between two classes in an ontology may not coincide with
some ontological distinctions (such as between {\em occurrent} and
{\em continuant}) that are deemed to be ``correct'' within the domain
\cite{Obrst2007}, but lead to undesired results in a similarity-based
analysis.

Finally, Widely used criteria for ontology development are the OBO
Foundry principles\footnote{Both accepted and proposed principles can
  be found on \url{http://obofoundry.org/crit.shtml}}. The accepted
criteria (as of 12 Feb 2012) include that ontologies must be (1) open
and freely available to all users, (2) that they are expressed using a
common syntax, (3) that they use unique URIs, (4) that they include
versioning information, (5) that their content is clearly delineated
and orthogonal to other ontologies, (6) that they contain natural
language definitions for all their terms, (7) that they define
relations based on patterns described in the OBO Relationship
Ontology, (8) that they are well-documented and (9) have multiple,
mutually independent users, (10) that they are developed
collaboratively while (11) only a single person is responsible for the
ontology, (12) that they follow ontology naming conventions, and (13)
that they are maintained in light of scientific advance. The majority
of these criteria (1-5, 9-13) are intrinsically social criteria;
although their discussion is outside the scope of the current article,
it must be emphasized that these criteria are highly valuable for
enabling wide access to the content of the ontologies within the OBO
Foundry, and therefore serve to enable scientific discourse about and
investigations into the ontologies and their content.  The remaining
criteria could be classified based on methods to demonstrate that they
are satisfied and based on the tasks which they aim to improve. For
example, while the inclusion of textual definitions (criterion 6) and
documentation (criterion 8) may improve comprehensibility of
ontologies, comprehensibility also depends on the quality of the
textual definitions and documentation; user-studies may be used to
evaluate and quantify the effect of these criteria, and compare them
against automated methods to generate textual definitions
\cite{Stevens2010}. Criterion (7), the use of relations that are
defined in the OBO Relationship Ontology \cite{Smith2005}, aims to
improve interoperability between ontologies through reuse of
relations. However, while relation {\em names} may be reused across
ontologies, it is not always guaranteed that they are reused in the
same {\em meaning}. To quantify whether criterion (7) succeeds in
enabling interoperability between ontologies, it would, for example,
be possible to combine two ontologies that both use relations from the
OBO Relationship Ontology, and evaluate whether or not they yield
desired inferences (i.e., a comparison of inferences against a gold
standard).


\section{Conclusions}
Our central position is that research results in {\em applied}
ontology should always be evaluated against a task for which they are
intended, i.e., the evaluation must be based on the behavior of the
whole system consisting of the ontology and the applications that are
based on it. Whether the research result is an ontology, or an
ontology design pattern, or a method to formulate particular phenomena
within a domain, the benefit it can bring to the domain cannot be
evaluated based on the research result alone; instead, any evaluation
criteria must evaluate the whole system consisting of the research
result and a task -- or a set of tasks -- to which the ontology-based
research is being applied.

Many of the applications and tasks in which ontologies play a role are
amenable to quantitative evaluation criteria. Quantitative measures
enable the objective comparison of research results and can play a
crucial role in the evaluation of research. We have reviewed several
common applications of applied ontology research in biomedicine, and
discussed potential quantitative evaluation measures for each of
them. 

These quantitative measures could be adopted in addition to already
established qualitative evaluation criteria, and they can also serve
to justify and refine existing qualitative measures.  For example,
while we have little doubt that qualitative measures such as formal
consistency and the absence of contradictory statements in an ontology
are useful and important quality criteria, we believe that many of
these qualitative criteria can be derived from underlying quantitative
measures of the performance of ontology-based research within a task:
consistency of ontologies is a useful criterion {\em because} many
applications of ontologies depend on consistency and {\em because}
consistent ontologies will often lead to better outcomes in whatever
application an ontology is being applied for.

Furthermore, with the application of quantitative measures, ontology
development methodologies can be evaluated with respect to how well
they ensure or improve the performance of research results in
particular tasks within a domain. More importantly, accepted
evaluation criteria for research results are the first step in
developing a research methodology for the field of applied ontology.
It was not our aim to establish such criteria for research in applied
ontology; instead, we believe that we, as a community of scientists
and scholars, must increase our efforts towards establishing such
evaluation criteria for research in applied ontology, based on which
we can derive a research methodology within our field.


\section*{Acknowledgements}
We are grateful to Prof. Heinrich Herre, who contributed to this work
over many years through discussions and critical remarks.

\bibliography{../bibtex/lc}

\end{document}